\documentclass{ifacconf}
\usepackage{array}
\usepackage{amsmath,amssymb,amsfonts}
\usepackage[utf8]{inputenc}
\usepackage{graphicx}      
\usepackage{algorithm}
\usepackage[noend]{algpseudocode}
\usepackage{booktabs}
\usepackage{natbib}        
\usepackage[absolute]{textpos}
\newcommand{\IFACCopyRightStatement}
{
\begin{textblock*}{230mm}(-30mm,23mm)  
\noindent
\footnotesize 
© 2024 the authors. This work has been accepted to IFAC for publication under a Creative Commons Licence CC-BY-NC-ND
\end{textblock*}
}

\begin{document}
\begin{frontmatter}
\IFACCopyRightStatement
\title{Hybrid Unsupervised Learning Strategy \\ for Monitoring Industrial Batch Processes}

\author[First]{Christian W. Frey} 

\address[First]{Fraunhofer Institute of Optronics, System Technologies and
Image Exploitation IOSB, Karlsruhe, Germany
\\ (e-mail: christian.frey@iosb.fraunhofer.de)}

\begin{abstract}
Industrial production processes, especially in the pharmaceutical industry, are complex systems that require continuous monitoring to ensure efficiency, product quality, and safety. This paper presents a hybrid unsupervised learning strategy (HULS) for monitoring complex industrial processes. Addressing the limitations of traditional Self-Organizing Maps (SOMs), especially in scenarios with unbalanced data sets and highly correlated process variables, HULS combines existing unsupervised learning techniques to address these challenges. To evaluate the performance of the HULS concept, comparative experiments are performed based on a laboratory batch process.\end{abstract}

 \begin{keyword} AI methods for FDI, condition monitoring, neural network, data-driven, unsupervised machine learning, self-organizing maps, industrial manufacturing process
\end{keyword}

\end{frontmatter}

\section{Introduction} \label{introduction}
Continuous monitoring of complex manufacturing processes involves the systematic surveillance, measurement, and analysis of various aspects of a production system to ensure efficient operation, maintain product quality, and comply with safety standards. Anomaly detection is a critical aspect of a comprehensive monitoring system and involves the identification of atypical patterns, events, or behaviors in datasets that significantly deviate from the norm or expected values. In the domain of manufacturing processes, multivariate datasets encompass correlated time-sequenced records of process values. Multivariate anomaly detection techniques can be divided into two main categories: statistical and machine learning (ML) methods (\citealp{Schmidl2022}). While production processes are characterized by high system order with complex interactions between multiple process values, machine learning (ML) methods are preferred - they can effectively learn and model complex, non-linear relationships (\citealp{Helwig2015}). Supervised ML methods require labeled training data (e.g., normal or abnormal behavior), while unsupervised methods do not. Unsupervised ML methods assume that the recorded data reflects the normal behavior of the process, thus reducing the cost of data preparation and labeling. Clustering-based ML techniques group similar data points into clusters - anomaly is measured by how well a data point fits into the identified clusters. An additional layer of analysis is added by interpreting the clusters as characteristic phases of a process (e.g., start, steady, transient, end). This can be useful when monitoring batch processes in the pharmaceutical industry, for example, where the duration and sequence of process phases (phase trajectory) are critical in terms of product quality. Self-organizing maps (SOMs), introduced by \cite{Kohonen2013}, have been successfully employed for process monitoring in various applications. While the approaches presented in \cite{Vermasvuori2002TheUO, Li2021AnomalySOM} focus on anomaly detection, the concepts presented in \cite{Birgelen2017UsingSM, Frey2008MonitoringOC} extend the utility of SOMs to identify, visualize, and monitor process phases. However, SOMs have certain limitations, especially when dealing with unbalanced training datasets and highly correlated process variables. To overcome these problems, a hybrid unsupervised learning strategy (HULS) have been presented by \cite{Frey2023AHU}. In order to evaluate the performance of the developed method, a comparative study based on a synthetic dataset was conducted. The experiments showed that the HULS concept is significantly superior, especially in the reliable identification of unknown process phases.
In this study, we explore the use of the HULS procedure for monitoring a laboratory batch process. Section~\ref{methodologies} outlines the basic methods used, while Section~\ref{sectionHULS} gives an overview of the HULS approach. Section~\ref{sectionexperiments} presents the experimental results, comparing the performance of HULS with SOM method. Finally, the paper is concluded in Section~\ref{sectionconclusion}.
\section{Methodologies} \label{methodologies}
\subsection{Self-Organizing Maps} \label{msoms}
SOMs are capable of mapping a high-dimensional input dataset $\mathbf{X}$ onto a low-dimensional grid while preserving the topological structure of the data. Usually structured in two layers, the input layer receives the input data points from the dataset $\mathbf{X}=\{ \mathbf{x}_1,...,\mathbf{x}_L  \}$. Each data point is represented as an vector with $\mathbf{x}_i \in \mathbb{R}^d \ \forall \ i$ where $d$ corresponds to the number of dataset features. The output layer is typically a 2D grid of size ${M\times N}$ with $P_{SOM} = MN$ neurons at position $\mathbf{v} \in \mathbb{N} = (1,...,M)\times(1,...,N)$. The Best Matching Unit (BMU) is the neuron with index $\mathbf{v}^*_i = \operatorname*{argmin}_\mathbf{v} \left\Vert \mathbf{x}_i - \mathbf{w}_\mathbf{v} \right\Vert$ whose weight vector $\mathbf{w}_\mathbf{v} \in \mathbb{R}^d \ \forall \ \mathbf{v}$ is the most similar to data point $\mathbf{x}_i$, with similarity typically measured by the Euclidean distance:
\begin{equation}
 e_{i\mathbf{v}} = \left\Vert \mathbf{x}_i - \mathbf{w}_\mathbf{v} \right\Vert. 
\end{equation}

In the $k$-th step of $K$ training epochs the weights of the BMU and its neighbors are updated by:
\begin{equation} \label{eq_weight_update}
  \mathbf{w}_\mathbf{v}(k+1) = \mathbf{w}_\mathbf{v}(k) + \alpha(k) h_{\mathbf{v^*}\mathbf{v}}(k) (\mathbf{x}_i - \mathbf{w}_\mathbf{v}), 
\end{equation}
where $\alpha(k)$ is the learning rate and $h_{\mathbf{v^*}\mathbf{v}}(k)$ is the neighborhood function. Both functions can be applied in several ways (\citealp{Kohonen2013}), however they should be decreasing with $k$. With $\alpha_0$ being the initial learning rate and $K$ the number of total learning epochs, $\alpha(k)$ is calculated by:
\begin{equation}
\alpha(k) = \alpha_0\exp \left(-\frac{k}{K}\right).
\end{equation}
The neighborhood function $h(k)$ defines the extend of influence of the BMU on its neighbors. The purpose is to adjust not only the BMU, but also the neighboring neurons, allowing the map to "self-organize" over time. The influence radius $\sigma(k)$ is usually a Gaussian function with initial influence radius $\sigma_0$:
\begin{equation}
h_{\mathbf{v^*}\mathbf{v}}(k) = \exp \left(-\frac{\left\Vert \mathbf{v}^*-\mathbf{v}\right\Vert}{\sigma_0 \sigma(k)}\right), \: \: \sigma(k) = \exp \left(-\frac{k}{K}\right).
\end{equation}
Because mapping quality of a trained SOM cannot be measured by a single energy function several metrics are available in the literature (\citealp{Erwin1992SelforganizingMO}). In this study, quantization error $E_Q$ and topographic error $E_T$ are chosen: 
\begin{center}\vspace{0pt}
\begin{minipage}[b]{.53\linewidth}\hfill
  \begin{equation} \label{eqEQ}
  E_Q = \frac{1}{L}\sum_{i=1}^L \left\Vert \mathbf{x}_i - \mathbf{w}_\mathbf{v^*} \right\Vert
  \end{equation}
\end{minipage}%
\begin{minipage}[b]{.47\linewidth}\hfill
  \begin{equation} \label{eqET}
   ,\: \text{and} \: \:E_T = \frac{1}{L}\sum_{i=1}^{L} \delta_i.
  \end{equation}
\end{minipage}%
\end{center}
The quantization error $E_Q$ measures the mean error between the data points $\mathbf{x}_i$ and the corresponding BMU weight vector $\mathbf{w}_\mathbf{v^*}$, while the topographic error $E_T$ measures how well spatial relationships are preserved when mapping $\mathbf{x}_i$ to the lattice. In a perfect SOM, the BMU and second BMU should be adjacent to reflect the topology of the input data ($\delta_i=0$). If they aren't, it's considered as a topographic error ($\delta_i=1$).
\subsection{Unified-Distance Matrix and Watershed Transformation} \label{mumatrix}
The unified-distance matrix (UM), presented by \cite{Ultsch2003MapsFT}, is valuable tool for visualizing and interpreting SOMs, providing insights into the underlying structure and clusters within the data. The base idea is to add a dimension to the SOM lattice, which reflects the summed distances between the individual neuron and its surrounding neighbors. With $\mathbf{v} = (m,n)$ be the position of the individual neuron and $\mathbf{p} = (m-1,...m+1)\times(n-1,...,n+1)$ be the position of the neighbors in the lattice the distance $U_\mathbf{v}$ is calculated by:
\begin{equation}
U_\mathbf{v} = \frac{1}{8}\sum_{p \neq v} \left\Vert \mathbf{w}_\mathbf{v} - \mathbf{w}_\mathbf{p} \right\Vert.
\end{equation}
As an example, Fig.~\ref{fig_soms} visualizes the UMs of a trained SOMs as 3D plot. In the context of monitoring production processes the valleys can be interpreted as distinct process phases while ridges mark transitions between the phases (\citealp{Frey2008MonitoringOC}). To automatically segment the UM, clustering algorithms, such as the watershed transformation (WT) can be used (\citealp{Vincent1991WatershedsID}). Output of WT is matrix $\mathbf{T}$ of size size ${M\times N}$. With a given margin $\phi$ the WT maps each element $T_\mathbf{v}$ to a cluster $c$, where $C$ is the set of identified clusters and $c \in \mathbb{N} = (0,1,...,|C|)$. Note that a neuron will be assigned to $c=0$ if the neuron weight $\mathbf{w}_\mathbf{v}$ has never been updated by (3). Algorithm~\ref{alg:som} summarizes the procedure of SOM training and clustering.
\begin{algorithm}
\caption{SOM training and clustering}\label{alg:som}
\begin{tabular}{ wl{4em} l}
 \textbf{Input:} & Initialized SOM, training data $\mathbf{X}$, epochs $K$,\cr 
 & learning rate $\alpha_0$, influence radius $\sigma_0$,\cr
 & watershed margin $\phi$\cr
 \end{tabular}
\begin{tabular}{ wl{4em} l}
\textbf{Output:} & SOM, UM $\mathbf{U}$, WT $\mathbf{T}$, clusters $\mathbf{C}$ \\
\end{tabular}
\hrule
\vspace{4pt}
\begin{tabular}{ wl{4em} l}
\textbf{Procedure:}  $SOM(\mathbf{X})$ \\
\end{tabular}
\begin{algorithmic}[1]
\For {epoch $k  <  K$}
\For {data point $\mathbf{x}_i \in X$} 
    \State Compute BMU $\mathbf{v}^{*}_i(k)$ (2)
    \State Adjust BMU and neighboring neurons (3)    
\EndFor
\EndFor
\State Compute UM $\mathbf{U}$ (7), WT $\mathbf{T}$
\end{algorithmic}
\end{algorithm}

\subsection{Instantaneous Topological Map} \label{mitm}
SOMs suffer from severe difficulties when the training data is unbalanced or when the features are highly correlated, which is typical in industrial manufacturing processes. There are several approaches to overcome these problems (\citealp{Ruppert2004TheEO}), one of which is the Instantaneous Topological Map (ITM) developed by \cite{Jockusch1999AnIT}. An ITM is also structured in two layers, where the input layer receives the data points $\mathbf{x}_i \in \mathbb{R}^d \ \forall \ i$ from the dataset $\mathbf{X}=\{ \mathbf{x}_1,...,\mathbf{x}_L  \}$. The output layer holds the neurons with an associated weight vector $\mathbf{s}_r \in \mathbb{R}^d \ \forall \ r$ with a total of $P_{ITM}$ neurons and a set of edges (neighboring neurons) $Q_j$. An  ITM has no fixed number of neurons $P_{ITM}$ and the creation of neurons is based on the relation between BMU $j^{*}$ and the second BMU $j^{**}$, which is determined e.g. by the Euclidean distance:
\begin{center}\vspace{0pt}
\begin{minipage}[b]{.47\linewidth}\hfill
  \begin{equation}
  j^*_i = \operatorname*{argmin}_\mathbf{j} \left\Vert \mathbf{x}_i - \mathbf{s}_j \right\Vert
  \end{equation}
\end{minipage}%
\begin{minipage}[b]{.53\linewidth}\hfill
  \begin{equation}
   ,  j^{**}_i = \operatorname*{argmin}_\mathbf{j \neq j^*} \left\Vert \mathbf{x}_i - \mathbf{s}_j \right\Vert.
  \end{equation}
\end{minipage}%
\end{center}
According to Algorithm~\ref{alg:itm}, ITM is an incremental procedure that updates the map instantaneously by sequentially processing incoming data points $\mathbf{x}_i$. In comparison to SOMs, the ITM method does not project input data onto a predefined lattice with a fixed number of neurons. Instead, ITM maps the dataset to a flexible number $P_{ITM}$ of neurons, with the topological relationships represented by the edges between neurons. The length of these connections indicates the topological proximity of the neurons involved. The ITM mapping sensitivity is regulated by the threshold parameter $\beta$, which corresponds to the radius of a sphere surrounding a neuron. A new neuron is created whether or not data sample falls within this sphere - the smaller the $\beta$ value, the more $P_{ITM}$ neurons are created: 
\begin{equation}
\beta \to 0: \: P_{ITM} \to L.
\end{equation}
This property of an ITM can also be interpreted in the context of resampling \cite{He2009LearningFI}, where the set of neuron weight vectors $\mathbf{S}$ reflect the resampled dataset $\mathbf{X}$. To assess the quality of the ITM mapping, the quantification error $E_Q$ given in (\ref{eqEQ}) can be employed, by using the BMU weight vectors $\mathbf{s}_{j^*}$.
\begin{algorithm}
\caption{ITM map generation}\label{alg:itm}
\begin{tabular}{ wl{4em} l}
 \textbf{Input:} & Initialized ITM, training data $\mathbf{X}$, threshold $\beta$ 
\end{tabular}
\begin{tabular}{ wl{4em} l}
\textbf{Output:} & ITM, set of weight vectors $\mathbf{S}$
\end{tabular}
\hrule
\vspace{4pt}
\begin{tabular}{ wl{4em} l}
\textbf{Procedure:} $ITM(\mathbf{X})$ \\
\end{tabular}
\begin{algorithmic}[1]
\For {data point $\mathbf{x}_i \in X$} 
    \State Compute BMU $j^{*}$ and second BMU $j^{**}$ (10)
    \If{$j^{*}$ and $j^{**}$ not connected} {add edge to $Q_{j^{*}}$}
    \EndIf    
    \For {neurons in $Q_{j^{*}} \in Q_{j^{**}}$}
        \If{$Q_{j^{*}}$ not Delaunay edge} {Remove $Q_{j^{*}}$} {}
        \EndIf
        \If{$j^{*}$ has no more edges} {Remove $j^{*}$} \EndIf
    \EndFor
    \If{$\left\Vert\mathbf{x}_i-\mathbf{s}_{j^{*}}\right\Vert>\beta \land   \left\Vert\mathbf{s}_{j^{*}}-\mathbf{s}_{j^{**}}\right\Vert < 
        \left\Vert\mathbf{x}_i-\mathbf{s}_{j^{*}}\right\Vert$} 
    \State {Add new neuron with $\mathbf{s}_{P+1} = \mathbf{x}_i$}
    \State {Add new edge between $\mathbf{s}_{j^{*}}$ and $\mathbf{s}_{P+1}$} to $Q_{j^{*}}$
    \State Compute BMU $j^{*}$ and second BMU $j^{**}$ (10)
    \If{$\left\Vert\mathbf{s}_{j^{*}}-\mathbf{s}_{j^{**}}\right\Vert < 0.5 \beta$} {Remove $j^{**}$} {}
    \EndIf
    {}
    \EndIf
\EndFor
\end{algorithmic}
\end{algorithm}
\section{Hybrid Unsupervised Learning Strategy} \label{sectionHULS}
As highlighted in section~\ref{introduction}, effective process monitoring involves several key aspects, including identifying unknown process phases, tracking the sequence of phases, the duration of individual phases, and detecting any anomalies that may occur within phases. Industrial applications, as referenced in sources like \cite{Frey2008MonitoringOC, Vermasvuori2002TheUO}, the capability of SOMs in detecting anomalies, identifying, and visualizing process phases is highly beneficial for the comprehensive monitoring and understanding of technical processes. From a data analysis perspective, this requires clustering and anomaly detection techniques. SOMs are able to perform both tasks simultaneously by discovering clusters within the training data and detecting any anomalies that do not conform to these clusters. \\
However, practical application faces several challenges mainly because SOMs are sensitive to the inherent properties of the training data. This sensitivity is particularly pronounced in cases where there is a strong correlation between different data features \cite{Ruppert2004TheEO}, as well as in situations where the dataset is unbalanced  \cite{He2009LearningFI}. SOMs, rely on the assumption that the features in the training data set are statistically uncorrelated (\citealp{Ruppert2004TheEO}). This premise may not hold true in the context of production systems, where the measured process values are time series that tend to be highly correlated. For instance, both temperature and pressure in a pharmaceutical reactor may be affected by the flow rate of the input materials. As a consequence, the SOM algorithm may struggle to effectively learn the true underlying data structure, lead to a biased map representation of the data structure and an increased topographic error $E_T$. Furthermore, dealing with an unbalanced training dataset, where certain classes are over- or underrepresented, can cause the SOM to prioritize the more common data points, potentially resulting in sub-optimal clustering with potentially suppressed clusters (\citealp{He2009LearningFI}). This aspect is crucial in exploratory process analysis, particularly when investigating unknown process phases. For example, batch production processes in the pharmaceutical industry are typically characterized by long-lasting stationary phases and brief transitions between them, resulting in an unbalanced training dataset. Prevalent strategy for managing unbalanced datasets is resampling, which involves either oversampling the minority class, undersampling the majority class, or both, to create a balanced dataset \cite{Krawczyk2016LearningFI}. Nonetheless, resampling methodologies are not easily applicable to the current problem formulation, particularly when e.g. the number of process phases are unknown. \\
To address these challenges, in \cite{Frey2023AHU} a hybrid unsupervised learning approach is proposed. The HULS concept combines the capabilities of SOMs and ITM. While ITM is also a topology preserving neural network it can handle correlated training data efficiently and re-sample the training data in an intuitive manner (\citealp{Barreto2003SelfOrganizingFM}). The output of an ITM is a set of neurons with associated weight vectors $\mathbf{s}_j$ and associated edges $Q_j$. However, unlike SOMs, ITMs do not map the input dataset $\mathbf{X}$ onto a 2D lattice and thus lacks an effective and intuitive clustering mechanism similar to e.g. UM based techniques. This is a key aspect considering the challenge of identifying and monitoring unknown process phases in an industrial manufacturing process. Therefore the proposed HULS concept for monitoring complex industrial manufacturing processes is based on the following idea: In the first step, the ITM algorithm is applied on the training dataset $\mathbf{X}$. Based on the threshold parameter $\beta$ of the ITM procedure the neuron generation is controlled, the resulting ITM neuron weights $\mathbf{s}$ can be understood as the resampled dataset of $\mathbf{X}$. In the next step, this set of ITM weight vectors $\mathbf{S}$ is then used as training dataset for the SOM procedure. Based on the trained SOM model the subsequent steps of UM transformation and WT segmentation can be performed.  Algorithm~\ref{alg:hybrid} summarizes the proposed HULS learning strategy.
\begin{algorithm}
\caption{Hybrid Unsupervised Learning Strategy}\label{alg:hybrid}
\begin{tabular}{ wl{4em} l}
 \textbf{Input:} & Training data $\mathbf{X}$, \cr
 & initialized ITM, threshold $\beta$,\cr 
 & initialized SOM, epochs $K$, learn rate $\alpha_0$, \cr 
 & infl. radius $\sigma_0$, watershed margin $\phi$\cr 
\end{tabular}
\begin{tabular}{ wl{4em} l}
\textbf{Output:} & ITM, SOM, UM, WT
\end{tabular}
\hrule
\vspace{4pt}
\begin{tabular}{ wl{4em} l}
\textbf{Pipeline:} & $HULS(\mathbf{X})$ \\
\end{tabular}
\begin{algorithmic}[1]
\State $ITM(\mathbf{X})$ (Algorithm~\ref{alg:itm})
\State $SOM(\mathbf{S})$ (Algorithm~\ref{alg:som})
\end{algorithmic}
\end{algorithm}
\section{Experiments} \label{sectionexperiments}
Industrial manufacturing processes, especially in the chemical and pharmaceutical industries, are characterized by high system order with a high degree of correlation between the different process values. These processes typically have distinct process phases, each with specific operating parameters and goals. However, access to real-world benchmark data sets from these industries is challenging due to their complex nature and confidentiality concerns. Considering these premises, the following comparative experiments will utilize a laboratory batch process installed in our lab for mixing and blending pharmaceutical pre-products. 
\subsection{Laboratory batch process}
\begin{figure}[htbp]
\centerline{\includegraphics[width=0.95\linewidth]{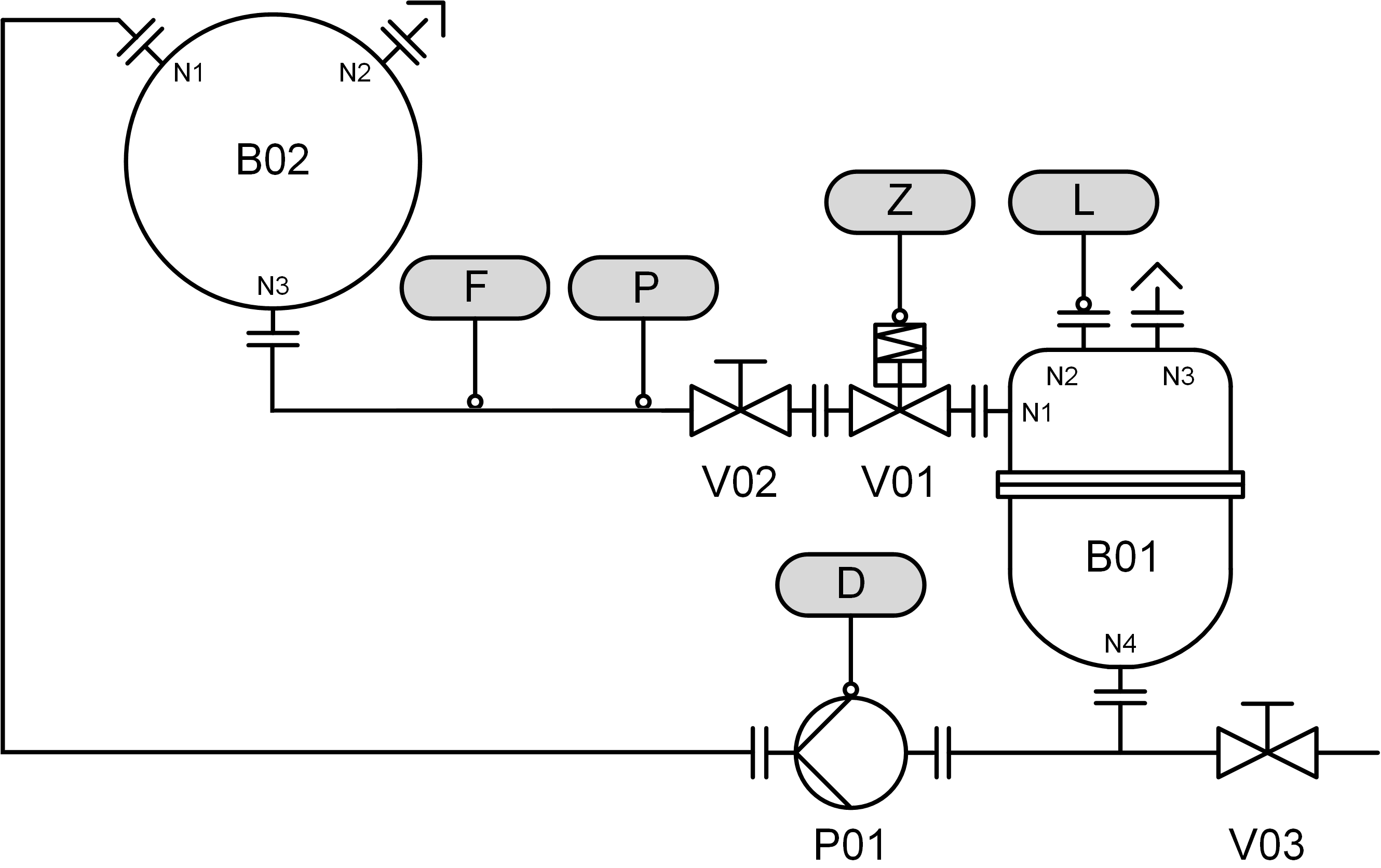}}
\caption{ Flow chart of the simplified laboratory batch process with measured values flow (F), pressure (P), level (L), and control values pump speed (D), and valve position (Z)}
\label{fig_laboratory batch process structure}
\end{figure}

Fig.~\ref{fig_laboratory batch process structure} shows the flow chart the considered laboratory batch process. The simplified process includes two liquid tanks, with tank B02 functioning as a supply reservoir for the process liquid. The outlet of tank B02 is equipped with a pressure sensor (P) for estimating the fill level, a flow meter (F), and an electro-pneumatic valve (Z) for regulating the flow rate of the liquids feed into the reaction tank B01. The B01 fill level is measured by an ultrasonic level gauge (L). Once B02 is filled to a defined level and after a product-specific reaction time, the liquid is transferred to a succeeding process unit using pump P01 with speed control (D) to adjust the flow rate. It should be noted that for the experiments, to illustrate a batch process behavior, the pre-product is returned to the supply tank B02. The sensors and actuators of the process are connected via an industrial fieldbus, which enables real-time data acquisition.
\subsection{Data acquisition}
\begin{figure}[htbp]
\centerline{\includegraphics[width=1\linewidth]{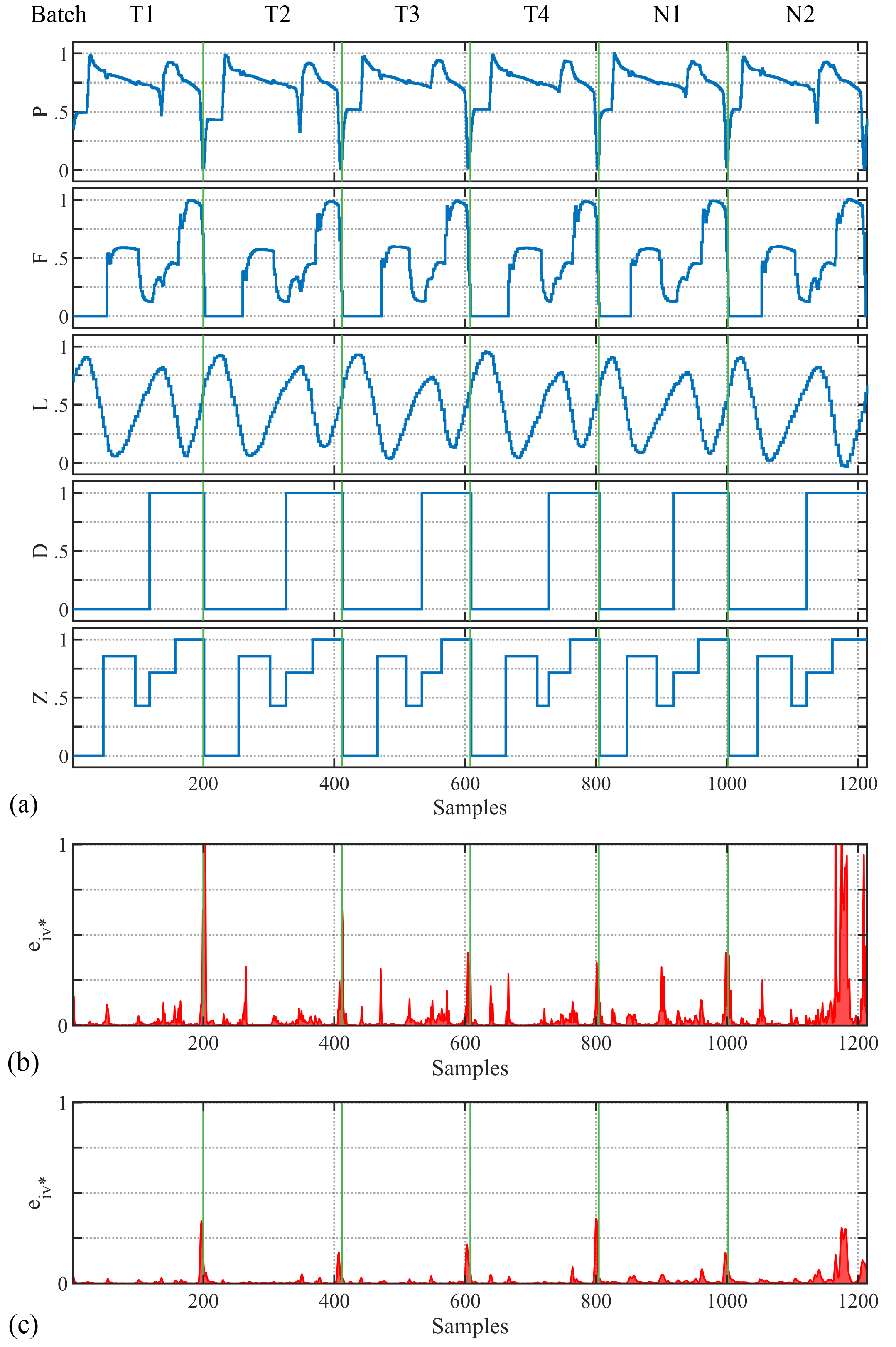}}
\caption{ (a) Recorded training and verification dataset. (b) Resulting quantification error $e_{i\mathbf{v^*}}$ for standard SOM model and (c) for proposed HULS procedure.}
\label{fig_trainingdataset}
\end{figure}

Data acquisition is the critical process of collecting and preparing diverse datasets, that are essential for training models to accurately reflect real-world scenarios. The model training phase uses these data to train, while the subsequent validation step tests the model on a separate dataset to ensure its ability to generalize and perform effectively on new, unseen data. For the following experiments it is assumed, that the training and validation datasets reflect the normal behavior of the process. With respect to this assumption, a training and a validation dataset, visualized in Fig.~\ref{fig_trainingdataset} (a), have been recorded. The multivariate dataset consist of the five process values (P,F,L,D,Z). These values are recorded over six batches, with batches T1 through T4 used for training and batches N1 and N2 used for validation. The length of the batches varies from 200 to 220 samples per batch. Note that the process values shown are normalized and the time reference is anonymized due to confidentiality restrictions. 
\subsection{Model training and validation} \label{sectionmodeltraining}
In order to make a valid comparison between the standard SOM and the proposed HULS algorithm, a standardized training setup was chosen for the experiments. Both algorithms are initialized with a symmetric map size of $P_{SOM}=67\times67$ neurons, with training parameters set to $K=1000$ iterations, $\alpha_0=0.02$ as initial learning rate, and $\sigma_0=2$ as initial influence radius. The additional HULS threshold parameter is set to $\beta=0.01$. Note that the map size is deliberately chosen to be large in order to highlight the different characteristics of the algorithms being compared.\\
 As a first result, Table~\ref{tab:table1} provides a comparison of the key quality measures for the two algorithms. The parameters are calculated based on the training batches T1 to T4. Comparing the quantization error $E_Q$ gives an idea of how well the two models perform in covering the training dataset. The proposed HULS method significantly outperforms the standard SOM. This phenomenon becomes clear when comparing Fig.~\ref{fig_trainingdataset} (b) and (c), which show the time courses of the Euclidean distance $e_{i\mathbf{v^*}}$ between the BMU and the training data samples $\mathbf{x}_i$. The increased $e_{i\mathbf{v^*}}$ for the standard SOM is particularly evident when fast dynamic transitions in the process behavior occur. This phenomenon is typical in real-world applications: training data samples that reflect the fast transition phases are underrepresented, resulting in an unbalanced dataset. In such situations, the learning algorithm of the standard SOM fails to adjust the weight vectors sufficiently to cover and accurately map the training data, resulting in increased error $e_{i\mathbf{v^*}}$. Furthermore, this effect becomes even more pronounced when considering the validation batches N1 and N2. The standard SOM does not generalize and perform effectively on new, unseen data when trained on an unbalanced dataset. In comparison the HULS procedure can handle an unbalanced dataset more efficiently, thus confirming its practical applicability and robustness for industrial process monitoring. 
\begin{table}
\centering
\caption{Mapping quality for standard SOM and HULS procedure}
\label{tab:table1}
\begin{tabular}{lcc}
\midrule
 & {Quantization Error $E_Q$} & {Topographic Error $E_T$} \\
\midrule
SOM & 1.078 & 10.38 $\%$ \\
HULS & 0.481 & 4,53 $\%$ \\
\midrule
\end{tabular}
\end{table}
\subsection{Discovering unknown process phases} \label{sectiondiscover}
\begin{figure}[htbp]
\centerline{\includegraphics[width=1\linewidth]{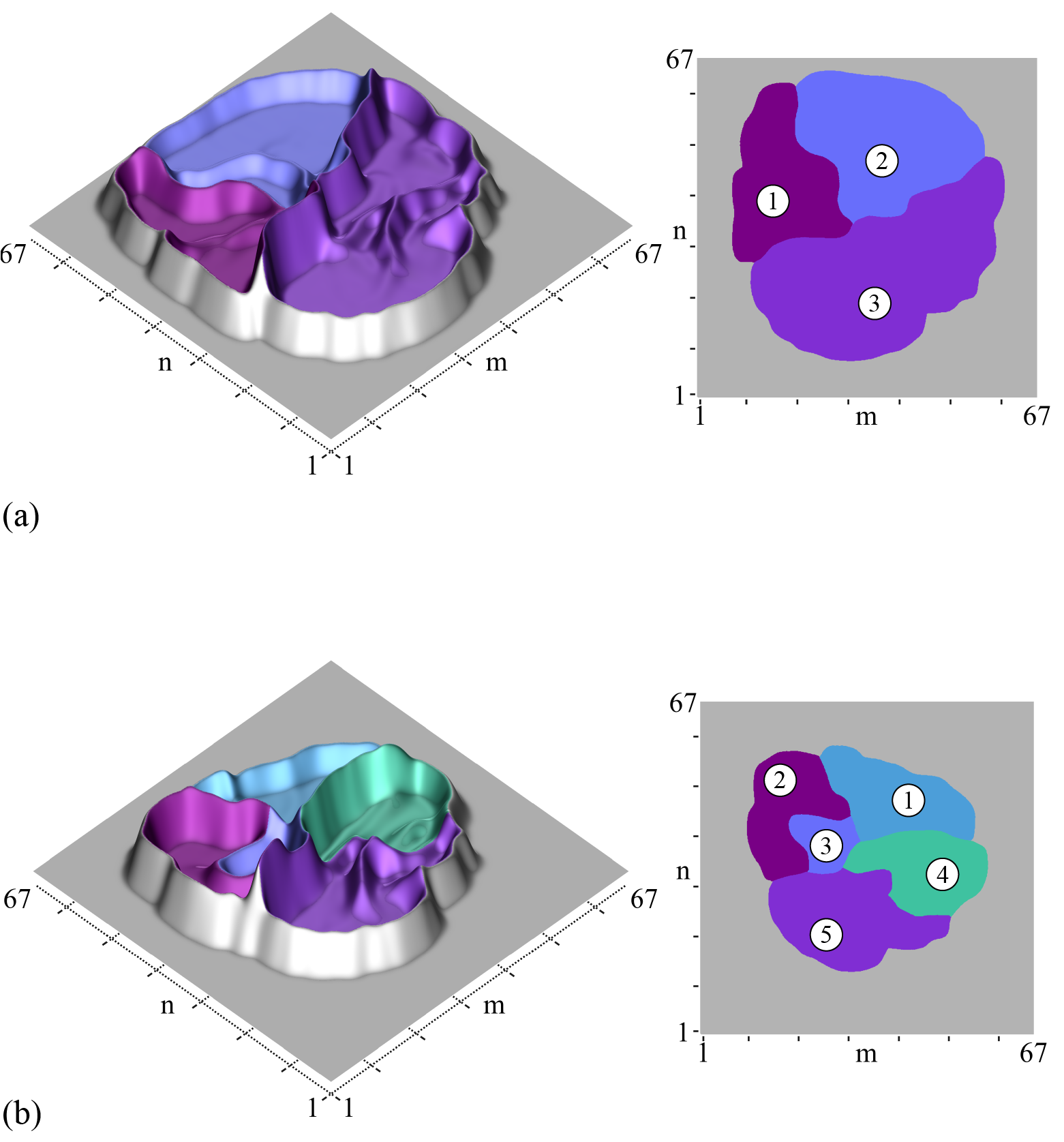}}
\caption{ 3D/2D visualization of UM and WT with corresponding WT clusters of the trained standard SOM (a) in comparison to the proposed HULS concept (b).}
\label{fig_soms}
\end{figure}

Holistic process monitoring, especially in the pharmaceutical industry, involves several essential elements such as identifying unknown process phases, tracking their progression, determining the duration of each phase, and detecting any anomalies. A prerequisite for robust process phase monitoring is the reliable clustering of the training dataset. As described in section~\ref{methodologies}, the use of SOMs with UM transformation and WT segmentation represents a robust approach for unsupervised clustering. As already discussed, the considered laboratory can be taken representative for a wide range of production processes: strong correlation among the process values and distinct process phases with fast transitions between them. However, the SOM algorithm exhibits considerable sensitivity to the inherent attributes of the training dataset, potentially leading to an increased topological error $E_T$, for example. As previously outlined, the $E_T$ measures the ability of the mapping algorithm to accurately model the topological dependencies within the dataset. As shown in Table~\ref{tab:table1}, the proposed HULS procedure again significantly outperforms the conventional SOM model. Evidently, a high topological error $E_T$ may cause the UM transformation to cluster the data insufficiently. This effect can be verified by examining the UM transformation and the WT segmentation.\\
For the following comparative evaluation, both the UM transformation and the WT segmentation algorithms were executed using a watershed margin of $\phi=10$'. The resulting UM with the identified clusters using the standard SOM algorithm is shown as a 3D and 2D plot in Fig.~\ref{fig_soms} (a). The standard SOM algorithm was able to identify three distinct process phases. This result contradicts the installed process flow control, with five distinct process phases implemented. In this simple example of the laboratory process, these phases can be read visually: by examining the time courses of the valve position (Z) in relation to the pump speed (D) in Fig.~\ref{fig_trainingdataset} (a), five dedicated combinations can be read for each batch. In contrast, as shown in Fig.~\ref{fig_soms} (b), the HULS concept successfully and reliably identifies the five process phases. When comparing the obtained results, the efficiency of the developed HULS concept becomes apparent. Even under challenging conditions, such as strong correlation between the process variables and the presence of an unbalanced learning dataset, the HULS method is able to reliable identify the process phases. Another aspect related to $E_T$ is the efficiency how the intrinsic structure of the data is mapped in the model. A direct comparison of the two UM visualizations in Fig.~\ref{fig_soms} shows that the HULS model is more compact, i.e. fewer neurons are needed to store the data structure. 
\subsection{Anomaly detection}
\begin{figure}[b]
\centerline{\includegraphics[width=1\linewidth]{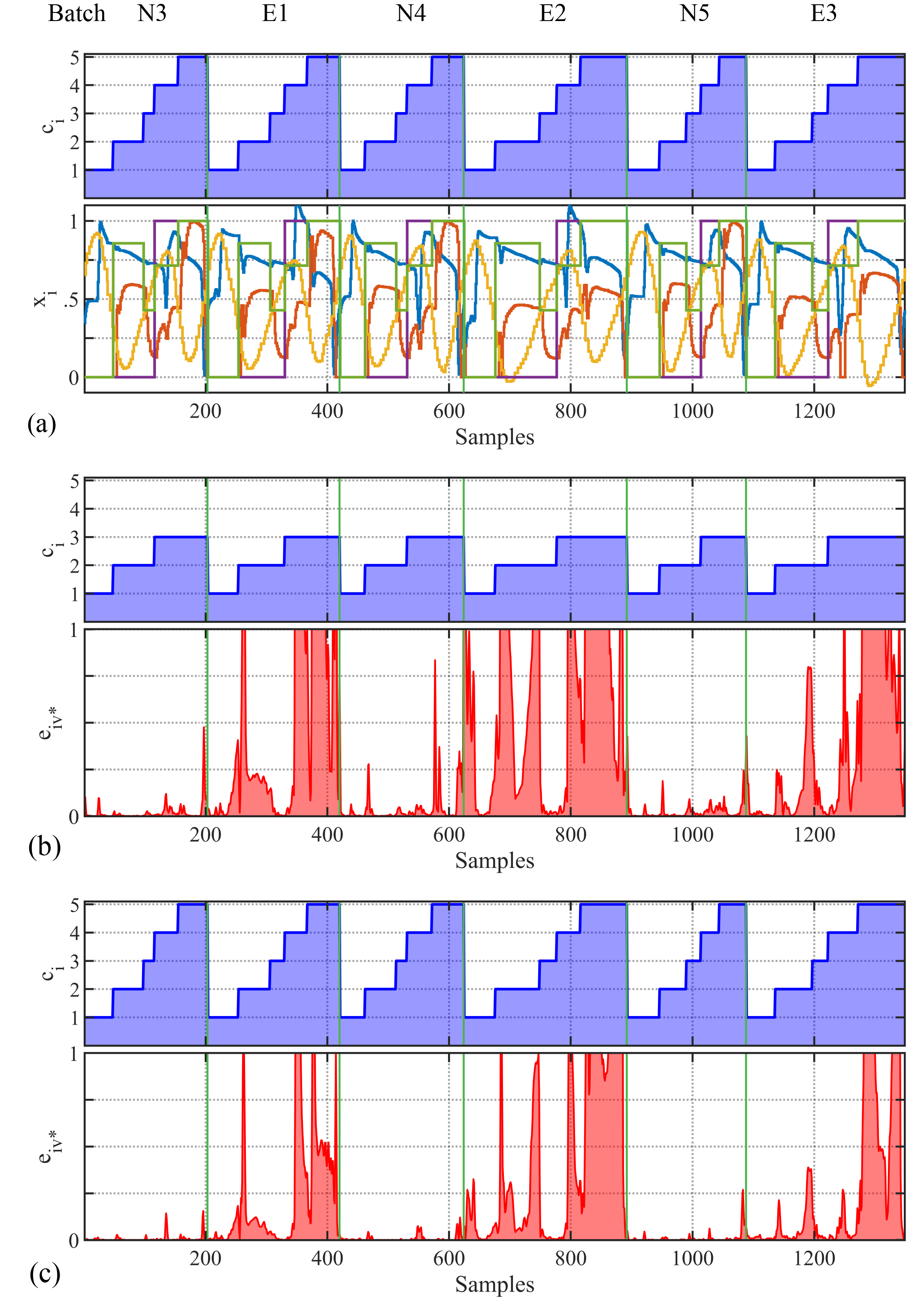}}
\caption{ (a) Recorded Batch sequence for evaluating anomaly detection performance. (b) Resulting quantification error $e_{i\mathbf{v^*}}$ and class assignment $c_i$ for standard SOM model and (c) for proposed HULS procedure.}
\label{fig_errordataset}
\end{figure}

Anomaly detection is a crucial aspect of a comprehensive monitoring system and involves identifying atypical patterns, events, or behaviors that significantly deviate from normal behaviour. For the comparative evaluation of the anomaly detection performance, another batch sequence was recorded. As shown in Fig.~\ref{fig_errordataset}(a) the sequence contains six batches, of which the three sequences E1, E2 and E3 deviate from the normal behavior. In batch E1 the venting of reservoir B02 (cf. Fig.~\ref{fig_laboratory batch process structure}) has been reduced, in batch E2 the flow cross-section has been decreased by valve V02 and in batch E3 the level gauge position in tank B01 has been shifted. Batches N3 to N5 reflect the normal behavior of the process. In Fig.~\ref{fig_errordataset} (b) and (c) the time courses of the Euclidean distance $e_{i\mathbf{v^*}}$ as well as the identified process phases $c_i$ are visualized. Focusing on batches N3, N4, and N5, where normal behavior is expected, shows that the standard SOM model is very sensitive to new, unseen data, resulting in a high $e_{i\mathbf{v^*}}$. This phenomenon, already described in section~\ref{sectionmodeltraining}, is problematic with respect to a real industrial application in the sense that false alarms can be generated (e.g. batch N4). \\
In terms of anomaly detection performance, both algorithms perform very well in terms of reliably detecting the anomalies in the batches E1, E2, and E3, but the HULS procedure still performs better in terms of recovering from the error state. Finally, regarding the process phase assignment $c_i$, it is clear that the standard SOM model cannot correctly identify the process phase because, as noted in section ~\ref{methodologies}, only three phases could be identified.
\section{Summary} \label{sectionconclusion}
The paper presents a hybrid unsupervised learning strategy (HULS) to improve monitoring in complex industrial processes. It addresses the limitations of SOMs in handling unbalanced and correlated data. To overcome these issues, HULS combines SOMs with the capabilities of ITMs in handling complex data scenarios. The HULS concept was comparatively tested on a laboratory batch process, demonstrating improved ability to identify process stages and detect anomalies. The strategy showed significant improvements over traditional SOMs, especially in industrial processes with strong correlations between the process values and pronounced process phases. The developed concept is currently being tested and evaluated in various industrial applications. The applications range from process industry, energy supply to monitoring of aggregates in the chemical industry. \\
\bibliography{ifacconf} 
\end{document}